\pdfoutput=1

\documentclass[11pt]{article}

\usepackage[final]{acl}

\usepackage{times}
\usepackage{latexsym}

\usepackage[T1]{fontenc}

\usepackage[utf8]{inputenc}

\usepackage{microtype}

\usepackage{inconsolata}

\usepackage{graphicx}

\usepackage{amsmath}
\usepackage{amssymb}
\usepackage{mathtools}
\usepackage{amsthm}
\usepackage{diagbox}
\usepackage{multirow}
\usepackage{enumitem}
\usepackage{booktabs}

\title{Reducing Distraction in Long-Context Language Models \\ by Focused Learning}

\author{
 \textbf{Zijun Wu\textsuperscript{\textdagger}\thanks{Work done during an internship at Amazon}},
 \textbf{Bingyuan Liu\textsuperscript{\textdaggerdbl}},
 \textbf{Ran Yan\textsuperscript{\textdaggerdbl}},
 \textbf{Lei Chen\textsuperscript{\textdaggerdbl}},
 \textbf{Thomas Delteil\textsuperscript{\textdaggerdbl}},
\\
 \textsuperscript{\textdagger}University of Alberta,
 \textsuperscript{\textdaggerdbl}AWS AI Labs,
\\
 \texttt{zijun4@ualberta.ca}\\
 \texttt{\{lbingy, yankran, chenzlei, tdelteil\}@amazon.com} \\
}

\begin{document}
\maketitle
\begin{abstract}
Recent advancements in Large Language Models (LLMs) have significantly enhanced their capacity to process long contexts. However, effectively utilizing this long context remains a challenge due to the issue of distraction, where irrelevant information dominates lengthy contexts, causing LLMs to lose focus on the most relevant segments. To address this, we propose a novel training method that enhances LLMs' ability to discern relevant information through a unique combination of retrieval-based data augmentation and contrastive learning. Specifically, during fine-tuning with long contexts, we employ a retriever to extract the most relevant segments, serving as augmented inputs. We then introduce an auxiliary contrastive learning objective to explicitly ensure that outputs from the original context and the retrieved sub-context are closely aligned. Extensive experiments on long single-document and multi-document QA benchmarks demonstrate the effectiveness of our proposed method.
\end{abstract}

\section{Introduction}

\begin{figure}[!t]
\centering
\includegraphics[width=0.48\textwidth]{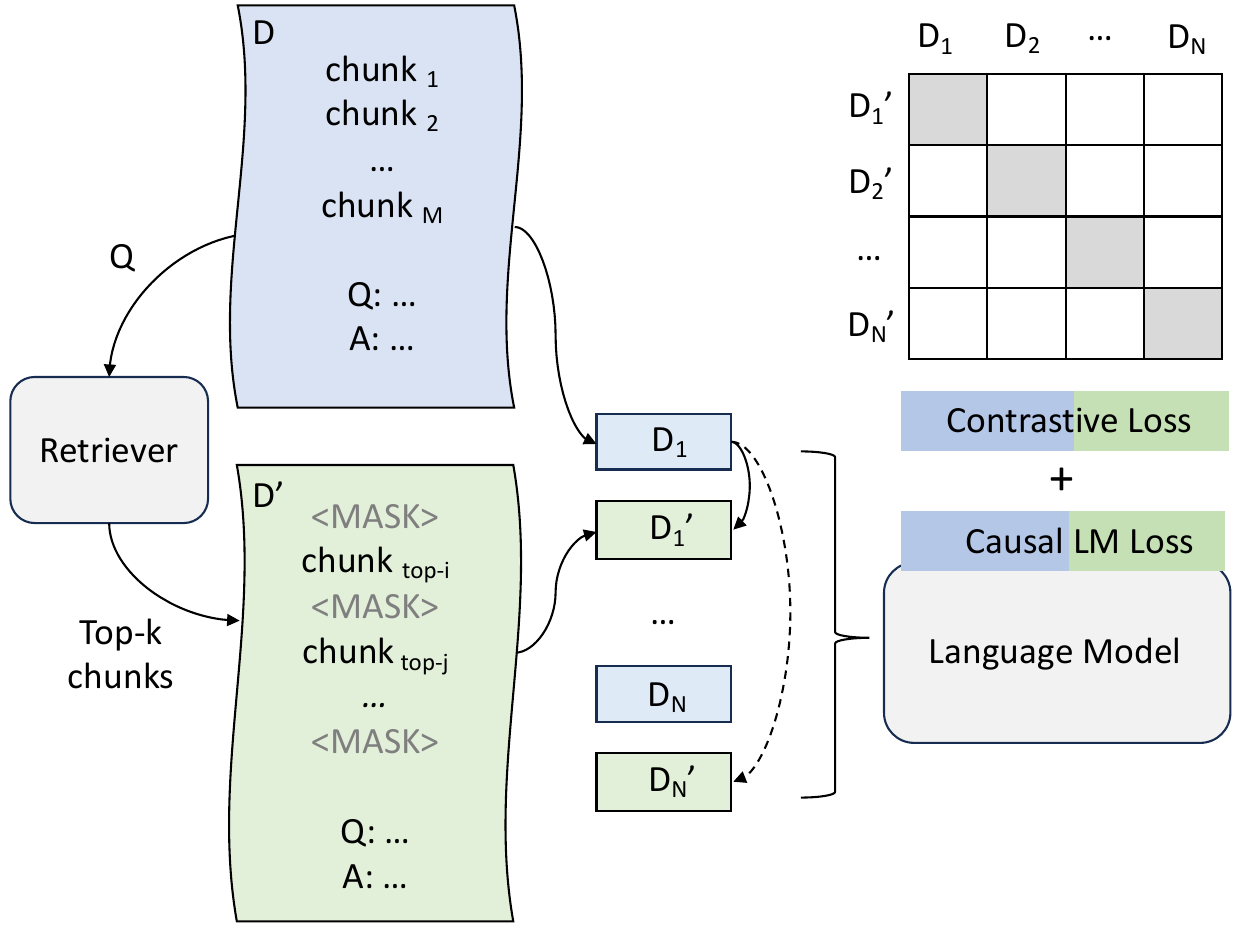}
\caption{Our method. \textbf{Retrieval-based data augmentation:} we filter out the distracting content from a document D' using a retriever, retaining only the top-k relevant chunks. The irrelevant portions are replaced with the $\texttt{<mask>}$ tokens. \textbf{Contrastive Training:} taking $D_1$ as an example, an augmented $D'_1$ is considered a positive pair with $D_1$ (solid line), whereas the augmented versions of other documents $D'_2, \cdots, D'_N$ serve as negative pairs (dashed line) for $D_1$.}
\label{fig: our method}
\end{figure}

Large language models (LLMs), such as the GPT series~\citep{NEURIPS2020_1457c0d6}, have established a new paradigm in natural language processing, showcasing exceptional versatility across various tasks~\citep{NEURIPS2020_1457c0d6, NEURIPS2022_9d560961}.
Efforts to enhance the contextual capabilities of LLMs have primarily focused on techniques like context extension fine-tuning~\cite{chen2023extending, chen2023longlora, ding2023longnet}, or retrieval augmented generation~\cite{lewis2020retrieval, xu2023retrieval, gao2024retrievalaugmented}. 
Despite these advancements, LLMs often struggle to effectively utilize extended contexts, frequently encountering the distraction issue~\citep{liu2023lost}. This problem arises when LLMs are easily distracted by irrelevant information within a long context.

The distraction issue presents a significant challenge in practical applications, especially in long-context question answering (QA) tasks~\citep{pang-etal-2022-quality, dasigi-etal-2021-dataset}. In these scenarios, the relevant information required to answer a question is often buried within lengthy texts. For example, the answer to a question may depend on a small segment of a long document. However, LLMs typically process input contexts holistically~\cite{NIPS2017_3f5ee243}, leading to an over-generalized distribution of attention across all tokens, which diminishes the model's ability to focus on the most relevant information.

One commonly considered solution is the utilization of a retriever during inference~\citep{pmlr-v119-guu20a, lewis2020retrieval}, where relevant information is extracted by the retriever as filtered input to enhance the LLMs' focus on essential sub-contexts~\cite{xu2023retrieval}.
However, crucial information may sometimes be excluded from the retrieved content due to the imperfections of retrievers. Such shortcomings in retrieval can lead to significant compounding errors or hallucinations in the generated responses~\citep{pmlr-v202-shi23a, liu2023lost}. 

In this study, we propose a novel training method to enhance long-context LLMs' inherent ability to focus on the relevant segments related to a specific question. Our technique integrates the ``focusing ability'' of a retriever with relatively shorter context length, into long-context LLMs through retrieval-based data augmentation and contrastive learning. Our approach eliminates the need for a separate retriever during inference, effectively addressing the issue of distraction. 

As shown in Figure~\ref{fig: our method}, our method contains two key ingredients: 1) \textbf{Retrieval-based data augmentation:} For each example, we generate an augmented input by retaining only the top-k retrieved segments associated with the question, masking irrelevant information with a special token. 2) \textbf{Contrastive learning:} We apply a contrastive learning~\citep{pmlr-v119-chen20j, radford2021learning} objective to enforce closer sequence representations of the original and its retrieval-augmented sample. This approach leverages the semantic equivalence of the retrieval-augmented sample to the original long context given the specific question, guiding the model to concentrate on the most relevant sub-context of a long input.

We validate our method using the \textit{Mistral}-7B model~\citep{jiang2023mistral}, employing low-rank adaptation (LoRA)~\citep{hu2022lora} for efficient fine-tuning.
Comprehensive results on two long single-document QA tasks (i.e., Qasper~\citep{dasigi-etal-2021-dataset} and QuALITY~\citep{pang-etal-2022-quality}) and a long multi-document QA task~\citep{liu2023lost} demonstrate that our method, with just a few hundred fine-tuning steps, significantly reduces distraction-induced errors, outperforming both standard training methods and retrieval-augmented inference techniques.

\section{Related work}
\textbf{Long Context LLMs.}
Recent efforts to extend the context window size of language models have focused on various approaches. One of the earliest directions is to replace dense attention with sparse attention to decrease the computational complexity brought by the long context input~\citep{child2019generating}, enabling models to be pre-trained on longer contexts~\citep{jiang2023mistral}. Another approach involves interpolating or extrapolating relative positional encodings~\citep{press2021train, su2024roformer}, showing that it is possible to make inferences with context lengths surpassing the pre-trained limit with minimal performance loss. Parallel research has investigated methods that do not necessitate model length extension, such as compressing input context by filtering with the concept of self-information~\citep{li-etal-2023-compressing} or using an off-the-shelf summarization model to shorten the context~\citep{fei2023extending}.

\textbf{Distraction Issues.}
Several studies have highlighted that language models are prone to distraction. \citet{pmlr-v202-shi23a} systematically evaluated the performance of various LLMs when irrelevant information is injected into their context for reasoning tasks, showing that irrelevant content distracts the model and leads to degraded performance.
\citet{liu2023lost} revealed a limitation in models with the capacity for long contexts: they often fail to fully utilize the context window, particularly when key evidence is in the middle, a phenomenon termed ``lost in the middle.'' 
To mitigate these distraction issues, various strategies have been proposed, such as instructing the model to ignore irrelevant content~\citep{pmlr-v202-shi23a} or introducing indicators for relevant content and prompt engineering the model to focus on these indicators~\citep{he2023never}. Another method employs a retrieval mechanism where relevant content is selected by a retriever and presented as a filtered input context~\citep{xu2023retrieval}. However, such inference-time retrieval methods can lead to a potential loss of global context and are sensitive to the granularity of the selected chunks.

\textbf{Retrieval-Augmented Generation (RAG).}
The integration of retrieval models with language models, known as retrieval-augmented generation (RAG), addresses the challenge of language models' limited access to updated knowledge by utilizing an external knowledge base~\citep{lewis2020retrieval, shi2023replug, asai2023self}. In this paradigm, language models and retrieval models undergo joint training to optimize their collaboration effectively. 
\citet{xu2023retrieval} concatenate a few selected chunks from a lengthy context using a retriever at inference time, which often risks losing critical global information. In contrast, our model embeds retrieval capabilities implicitly into its weights. This integration enables our model to maintain a holistic view of the entire context while selectively focusing on the most relevant content.

\textbf{Contrastive Learning on LLMs.}
Contrastive learning has shown its effectiveness for text generation models~\citep{lee2021contrastive, an2022cont, su2022a, jain-etal-2023-contraclm}. These methods conduct contrastive learning on encoder-decoder transformers to improve general language modeling tasks. \citet{caciularu-etal-2022-long} demonstrated that contrastive training on encoder-decoder transformers enables the model to differentiate relevant information from a long document. \citet{su2022a} and \citet{jain-etal-2023-contraclm} showed that shaping the last layer representation of the decoder-only model through contrastive learning can improve generation performance. To the best of our knowledge, we are the first to show that contrastive learning can help decoder-only LLMs focus better on the relevant content of a long input context.

\section{Approach}
In this section, we introduce our proposed training method to enhance LLMs' intrinsic ability to effectively utilize long contexts.

We start by discussing our data augmentation approach via retrieval to filter out irrelevant information (\S\ref{section: Augmenting Training Data for Filtering}). Next, we describe the causal language modeling (CLM) applied to both original and augmented data sequences (\S\ref{section: Causal Language modeling}), which provides the representations for the subsequent contrastive learning step (\S\ref{section: Training Contrastively for Focusing}). To efficiently manage memory usage while finetuning an LLM, we incorporate low-rank adaptation (LoRA) into our training objectives (\S\ref{section: Efficient Fintuning for Long Context}). 

\subsection{Retrieval-based Data Augmentation}
\label{section: Augmenting Training Data for Filtering}
We augment training samples with long context by employing a retrieval mechanism to filter out irrelevant or low-relevance information that is not directly needed for specific question-answering tasks.

While language models possess a holistic view of the entire context, they can be distracted by massive amounts of irrelevant content within a long context~\citep{liu2023lost}, leading to a loss of focus on relevant facts. 
Utilizing a dense retriever has proven effective during the model's inference phase~\citep{xu2023retrieval}, where retrieved sub-contexts are concatenated and serve as inputs for LLMs with a reduced context length, potentially minimizing distraction issues. However, this retrieval augmented method at inference time considers only the local context determined by the retriever, which may neglect global information and be sensitive to retrieval quality, leading to unrecoverable inaccuracies in generation.

Different from the inference-time methods, we employ a dense retriever to filter relevant context exclusively during training. Our intuition is that the retrieved content from a long context can provide useful supervision to teach the model where to focus. Specifically, each training sample, denoted as $x$, consists of a question $q$, an answer $a$, and a long context $D$. The context is heuristically divided into several chunks, denoted as $c_i$, where $i=1, \cdots, M$, and $M$ is the total number of chunks for the long context. The granularity of chunks can vary from a sentence to a paragraph or a fixed number of tokens.

We then utilize a state-of-the-art dense retriever~\citep{karpukhin-etal-2020-dense, izacard2022unsupervised}, an encoder model optimized to provide embeddings that exhibit high semantic similarity for pairs of related inputs. Initially, we encode the question and chunks as follows:
\begin{align}
    \mathbf{q} = \operatorname{Encoder}(q) , \quad \mathbf{c}_i = \operatorname{Encoder}(c_i)
\end{align}
We then obtain the relevance scores $S_i$ between the question and chunk $c_i$ based on the cosine similarity of their embeddings. 
\begin{align}\label{eqn: similarity score}
    S_i = \operatorname{sim}(\mathbf{c}_i, \mathbf{q})
\end{align}
Based on the relevance scores $S_i$, we select the in-context chunks that have the top-k $S_i$ as the filtered context. The remaining chunks are treated as distractors and masked using special \texttt{<mask>} tokens~\citep{zhang2020pegasus}. The augmented filtered context is then appended with the original question $q$ and the answer $a$, forming the augmented paired sample $x'$.

It should be noted that certain datasets (e.g., Qasper) provide gold evidence, comprising a few sentences or paragraphs essential to answering a particular question. In such cases, the gold evidence is considered higher quality annotated retrieved content than that extracted by a retriever model. We argue that the effectiveness of instructing models to focus on relevant subsections correlates directly with the quality of the augmented sample $x'$. Our experiments demonstrate that utilizing gold evidence yields the best performance.

\subsection{Causal Language Modeling}
\label{section: Causal Language modeling}
We define $x = [w_1, w_2, \ldots, w_T]$ as the sequence of tokens from original training data, where $T$ represents the length. Conversely, $x' = [w'_1, w'_2, \ldots, w'_{T'}]$ denotes the augmented sequence generated from $x$, with its length denoted by $T'$. Notably, $T' << T$ since the augmented sequence $x'$ retains only the relevant content.

Our approach involves fine-tuning a language model using a Causal Language Modeling (CLM) objective applied to both $x$ and $x'$, which is shown as follows:
\begin{align} \label{eqn: clm loss}
\begin{split}
\mathcal{L}_{\text{CLM}} = -\sum_{i=1}^{N} \bigg[ \sum_{t=1}^{T} &\log P(w_{t+1}^i | w_{1:t}^i) \\
+& \sum_{t=1}^{T'} \log P(w_{t+1}^{i'} | w_{1:t}^{i'})\bigg] .
\end{split}
\end{align}
By fine-tuning the language model on $x$, the model learns to format the outputs specific to the task. Additionally, fine-tuning on $x'$ is essential for forming sequence representations that are critical for our contrastive learning approach (discussed in the next section) that compares the representations of two sequences.

\subsection{Contrastive Learning for Focus}
\label{section: Training Contrastively for Focusing}
We argue that the augmented training sample $x'$, generated by using the retriever, is semantically equivalent to the original lengthy $x$ because it includes essential content needed to  answer the question.
To leverage this equivalence, we employ contrastive learning to enforce the model to produce similar sequence representations for both inputs.
This approach implicitly guides the model to concentrate on the most relevant content while maintaining an awareness of the global context.

Let $\mathbf{h}$ and $\mathbf{h}'$ be the representations of $x$ and $x'$, obtained from the representations of the end-of-sequence (EOS) token\footnote{In the decoder-only transformer such as the GPT~\citep{radford2018improving} models considered in this study, the EOS token attends to all previous tokens of the sequence, it is thus suitable to be used as the representation of sequence~\citep{radford2018improving, pmlr-v139-radford21a}.} from the output layer of transformer~\cite{NIPS2017_3f5ee243}. We denote $I = \{1,2,\ldots,N\}$ and $I' = \{1',2',\ldots,N'\}$ as the indices of N instances of $x$ and $x'$, respectively. 

For a batch of N instances $I \cup I'$, the objective of the contrastive learning is to maximize the similarity between the representations of the original and the augmented inputs ($\mathbf{h}_i$ and $\mathbf{h}_i'$), while pushing apart the representations of all other pairs. More formally, the objective is to minimize the following:
\begin{align}\label{eqn: contra los}
\begin{split}
\mathcal{L}_{\text{Contra}} = -\sum_{i=1}^{N} \bigg[ &\log \frac{\exp(\operatorname{sim}(\mathbf{h}_i, \mathbf{h}_i') / \tau)}{\sum_{j=1}^{N} \exp(\operatorname{sim}(\mathbf{h}_i, \mathbf{h}_j') / \tau)} \\
+ &\log \frac{\exp(\operatorname{sim}( \mathbf{h}_i', \mathbf{h}_i) / \tau)}{\sum_{j=1}^{N} \exp(\operatorname{sim}(\mathbf{h}_i', \mathbf{h}_j) / \tau)} \bigg] .
\end{split}
\end{align}
where the $\operatorname{sim}$ function denotes the cosine similarity between two representations, and $\tau$ is a temperature parameter that scales the logits in the softmax. We follow \citet{pmlr-v139-radford21a} and set $\tau$ as a learnable parameter through back-propagation.

\subsection{Efficient Fintuning for Long Context}
\label{section: Efficient Fintuning for Long Context}
Training language models with the transformer architecture~\citep{NIPS2017_3f5ee243} requires substantial memory, as computational complexity increases quadratically with the length of the input sequence.  LongLoRA~\citep{chen2023longlora} demonstrates that LoRA fine-tuning can adapt LLMs to longer contexts without sacrificing performance. Using LoRA allows for more efficient fine-tuning by significantly reducing the number of trainable parameters, which decreases memory usage and accelerates training.

We follow the approach of LongLoRA~\citep{chen2023longlora}, which involves adding adaptation to the query, key, value, and output attention weights ($W_q, W_k, W_v, W_o$) and make the embedding layer and layer-normalization layers tunable. We fine-tune a language model using a combination of the CLM and contrastive learning objectives from Equations~\ref{eqn: clm loss} and~\ref{eqn: contra los}, as follows:
\begin{align} \label{eqn: total loss}
    \mathcal{L} = \mathcal{L}_{\text{CLM}} + \mathcal{L}_{\text{Contra}}.
\end{align}
Due to the learnable nature of the temperature parameter $\tau$ in Equation~\ref{eqn: contra los}, the contrastive loss $\mathcal{L}_{\text{Contra}}$ can dynamically adjust its scale with the main CLM loss, we thus weigh two losses equally.

\section{Experiment}
\subsection{Datasets}
We consider three popular question-answering benchmarks, including both single and multi-document settings, to evaluate our method. All these benchmarks involve long-context input, which may introduce potential distraction issues. The statistics are detailed in Appendix~\ref{appendix: datasets}.

\textbf{Qasper}~\citep{dasigi-etal-2021-dataset}
is a single-document question-answering dataset on academic papers, specifically in the domain of NLP. Each sample contains a long context of the paper, a question, and an answer. During inference, the model is required to generate an output based on a paper and a question. We exclude the ``Unanswerable'' questions for a fair comparison with the inference-time retrieval method. Performance is evaluated by calculating the F1 scores between the outputs and the gold answers.

\textbf{QuALITY}~\citep{pang-etal-2022-quality}
is a single-document question-answering dataset on books and articles. Unlike the Qasper dataset, QuALITY is a multiple-choice dataset where each sample has four options. The model is required to select the correct options or answers. The performance metric is the accuracy of the correctly chosen options.

\textbf{Natural Questions with distractors (NQd)}
is a synthetic multi-document dataset based on the Natural Questions~\citep{kwiatkowski-etal-2019-natural}, inspired by \citet{liu2023lost}. For each question, the retriever is used to obtain 50 candidate documents, within which one gold document is embedded. These documents are concatenated to form the long context input, with the position of the gold document being random. Furthermore, we follow \citet{liu2023lost} to place gold documents at different positions at test time for in-depth analysis. We evaluate the Exact Match (EM) scores between the outputs and the gold answers.

\subsection{Experimental Setup}
We opted for \textit{Mistral}-7B~\citep{jiang2023mistral} model, a decoder-only pre-trained LLM for our experiments.
Our choice is informed empirically by the finding that the \textit{Mistral} model does not need to conduct length extension~\citep{xiong2023effective} by the continual pre-training, because it has been pre-trained with 32k context length. It is much longer than the average context length of the datasets considered in this study. Therefore, it can be directly fine-tuned with long-context samples of interest at a rapidly adaptive pace. The training procedure and hyperparameters are detailed in Appendix~\ref{appendix: training procedure}.

Regarding the retriever, we adopted the widely used \textit{Contriever}~\citep{izacard2022unsupervised}, which is an encoder-only transformer model pre-trained for information retrieval. 
It should be noted that the context length of \textit{Contriever} is only 512, which is much shorter than the full context length of the input to the language model, and the chunk size should also be smaller than the retriever's context length to fully utilize its retrieval ability.

We include two intuitive baselines to compare with our proposed method. 
\begin{itemize}
    \item \textbf{Vanilla training:} This approach involves fine-tuning the \textit{Mistral}-7B model without our proposed data augmentation and contrastive learning techniques. During inference, the vanilla-trained model processes long-context inputs the same way as our method.
    \item \textbf{Inference-time retrieval:} Following \citet{xu2023retrieval}, we integrated a retrieval pipeline to the vanilla-trained model at inference. Specifically, different from our proposed training-time retrieval method, we use the same retriever to filter out distracting content of the input at inference time. Additionally, we utilize the retriever to re-rank the documents for the multi-document setting, as previous studies have shown that models may focus more on the content at the beginning or end of its context~\citep{xu2023retrieval, liu2023lost}.
\end{itemize}
These baselines allow us to evaluate the effectiveness of our method against standard fine-tuning and retrieval-augmented inference approaches.

\subsection{Evaluation on Single-Document Tasks}\label{sec: Single-document Results}

\begin{table}[!t]
\centering
\resizebox{0.49\textwidth}{!}{
\begin{tabular}{c | c  c  c  c c} 
\toprule
\multirow{2}{*}{\textbf{Method}} & \multirow{2}{*}{\textbf{\#}} & \multirow{2}{*}{\textbf{Context}} & \multirow{2}{*}{\textbf{Query}} & \multicolumn{2}{c}{\textbf{Datasets}}\\
\cmidrule(lr){5-6}
&  &  & & \textbf{Qasper}& \textbf{QuALITY}\\
\midrule
Vanilla & 1 & - &  - & 48.65 & 47.17 \\
\midrule
\multirow{3}{*}{\shortstack{Inference\\-time}} & 2 &Fixed & Q & 40.54 & 50.14 \\
& 3 & Sent & Q & 38.62 & 44.63 \\
& 4 & Gold & - & 47.59 & - \\
\midrule
\multirow{5}{*}{Ours} & 5 & \multirow{2}{*}{Fixed} & Q & 51.31 & 47.94 \\
& 6 & & A & 45.00 & 49.42 \\
\cmidrule(lr){2-6}
& 7 &\multirow{2}{*}{Sent} & Q & 50.60 & 48.56 \\
& 8 & & A & 45.99 & \textbf{51.39} \\
\cmidrule(lr){2-6}
& 9& Gold & - & \textbf{59.62} & - \\
\bottomrule
\end{tabular}}
\caption{Results for single document retrieval settings. The terms ``Fixed,'' ``Sent,'' and ``Gold'' in the context column refer to retrieval using fixed-size chunking, sentence-level chunking, and gold evidence, respectively. ``Q'' and ``A'' in the query column indicate whether the retrieval query was the question or the answer. Note that using answer as query is only applicable in our training-time retrieval augmentation method.}
\label{tab:main single doc results}
\end{table}

We first evaluate the models on the Qasper and QuALITY datasets, where the context for question answering is a single document. To extract the relevant content, we utilize a retriever to select the top-k chunks based on the similarity scores from Equation~\ref{eqn: similarity score}. We define two types of chunk granularity for the in-context retrieval:
\begin{itemize}
    \item \textbf{Sentence-level chunking:} Each sentence is treated as a chunk. The retriever focuses on finding chunks containing entities mentioned in the query, which may result in the loss of some global semantic information. We extract the top 20 sentences and preserve their original order.
    \item \textbf{Fixed-size chunking:} The method allows overlaps between chunks. For instance, we use a fixed size of 500 tokens (the maximum context length of \textit{Contriever} is 512) with a 50-token overlap between consecutive chunks. We extract the top 3 chunks and maintain their original order.
\end{itemize}

Our method uniquely incorporates answers for retrieval during the training data augmentation phase. By using both questions and answers, we enhance the retrieval quality of relevant content. This improves the model's ability to focus on the most relevant information, leading to better performance~\citep{pang-etal-2022-quality}.

\textbf{Better Focus by Learning with Retrieval.}
Table~\ref{tab:main single doc results} compares the performance on the Qasper and QuALITY datasets. Integrating retrieval with the vanilla-trained model during inference results in mixed performances across different datasets.
In particular, the performance on QuALITY improves from 47.17 to 50.14 (Lines 1 and~2) with fixed-size chunking granularity. However, it experiences a decline in sentence-level retrieval.
A comparable trend of performance degradation is also noted on the Qasper dataset for both chunk and sentence-level retrieval settings, but both methods lag behind the performance of the vanilla method, underperforming by a considerable margin.

In contrast, our best method outperforms both the Vanilla and inference-time retrieval methods across both benchmarks. Using questions as queries to augment training data results in a marginal increase for the fixed-size chunks method, from 47.17 to 47.94 on QuALITY (Lines 1 and ~5), and more notably, to 48.56 (Line 7) at the sentence level. Utilizing answers as queries further elevates performance, reaching 51.39 (Line 8) at the sentence level. On Qasper, our method achieves its best results when employing questions for retrieval, with a significant to 51.31 with fixed-size chunking granularity (Lines 5). However, using answers for retrieval results in a performance decrease. Our manual inspection of gold answers in Qasper's training set suggests that questions tend to be more specific than answers, indicating that questions might be more effective as retrieval queries.

\textbf{Best Focus by Learning with Gold Retrieval.}
The Qasper dataset includes annotated evidence for each answer, which we consider as gold retrieval content, superior to that retrieved by the model. As seen in Table~\ref{tab:main single doc results}, performance significantly improves when augmented with this gold evidence compared to using the retrieval model, from 51.31 for fixed-size chunking and 50.60 at sentence-level chunking to 59.62 with gold evidence (Line 9). This enhancement corroborates the effectiveness of our approach, demonstrating that performance increases with the quality of retrieval during data augmentation. Conversely, a marginal decline in performance from the inference-time retrieval (line 4) is observed even with gold retrieval, from 48.65 to 47.59, further emphasizing the superiority of our approach.

Overall, unlike inference-time methods that heavily depend on a retriever to define the local context, our approach integrates the retriever's capabilities directly into the model's weights. This integration allows the model to maintain focus on relevant details without losing the global context. More importantly, our method eliminates the need for additional components during inference.

\subsection{Evaluation on Multi-Document Task}
We examine the effectiveness of our method in multi-document settings using the NQd dataset. Each sample in this dataset consists of one gold document and additional distracting documents, which are irrelevant for answering the posed questions. We did not use a retriever for data augmentation because each sample $x$ is synthesized by combining a known gold document with other distracting documents to fill the context length. Consequently, the context in the augmented document $x'$ during training always includes the gold document. 

The primary objectives of evaluating this dataset it to gain a deeper understanding of how models trained with and without our proposed method differ in their ability to focus on relevant documents. More importantly, assess how these models benefit from the use of a retriever at the inference stage. The results are shown in Table~\ref{tab:main nqd results}, where lines 1-3 represent the results from the vanilla-trained model, and lines 4-6 are from our method.

\begin{table}[!t]
\centering
\resizebox{0.49\textwidth}{!}{
\begin{tabular}{c l c c c c c c} 
\toprule
\multirow{2}{*}{\#} & \multirow{2}{*}{Methods} & \multicolumn{5}{c}{\#Documents} & \multirow{2}{*}{Avg.}\\
\cmidrule{3-7}
& & 10 & 20 & 30 & 40 & 50 &  \\ 
\midrule
1 &Vanilla & 51.6 & 45.2 & 45.2 & 42.0 & 40.0 & 44.6\\ 
2 &\, + Retrieval & 45.6 & 45.4 & 46.8 & 45.8 & 46.6 & 46.0 \\
3 &\, + Re-rank & 50.6 & 48.2 & 44.6 & 45.6 & 45.4 & 46.9\\
\midrule
4 & Ours   & 52.4 & 46.4  & \textbf{50.6} & 47.6 & 43.4 & 48.1 \\ 
5 &\, + Retrieval & 50.2 & 49.6 & 46.4 & 49.2 & 46.4 & 48.4 \\
6 &\, + Re-rank & \textbf{53.6} & \textbf{51.0} & 50.4 &  \textbf{53.6} & \textbf{51.6} & \textbf{52.0} \\
\bottomrule
\end{tabular}}
\caption{Results on the multi-document setting using the NQd dataset.}
\label{tab:main nqd results}
\end{table}

\textbf{Analysis of Tolerance in Higher Distraction.}
We established five different document lengths for inference. Intuitively, as the document length increases and with only one gold document present, the number of distracting documents also rises. This increment in distractors makes it increasingly challenging for the model to identify the truly relevant document to generate a correct answer. This trend is evident in lines 1 and 4 of our results, where no retriever is applied at inference. Both models, whether trained with our method or not, exhibit a decline in performance as the number of documents increases from 10 to 50, which aligns with the findings from~\citet{liu2023lost}. However, the model trained using our method consistently outperforms the vanilla-trained model across various document lengths.

Additionally, our model demonstrates a higher tolerance for distractions while maintaining superior performance. For example, when conducting inference with 40 documents, our model exhibits better performance compared to the vanilla-trained model working with 30 documents under less distraction (47.6 vs. 45.2). When examining the average performance, there is a notable improvement: 48.1 compared to 44.6. These results suggest that our method effectively aids the model in focusing on the relevant document despite the presence of numerous distractors.

\textbf{Inference-Time Retrieval Helps in High Distraction.} 
We applied a retriever to the vanilla-trained models, where only the top-ranked document returned by the retriever is selected and appended with the question as the input. As seen in lines 2 and 4 in Table~\ref{tab:main nqd results}, the model trained with our method still has an advantage over the inference-time retrieval method, with an averaged improvement across all document lengths from 46.0 to 48.1. 

Furthermore, we observe that when the context length is relatively short (i.e., the document number is 10), retrieving a relevant document using a retriever is not helpful and may even hurt performance. This suggests that the model's internal retrieval capabilities are superior to the external retriever's performance when the distraction level is low. Conversely, when the distraction level is high (i.e., the document number is over 20), utilizing retrieval at inference time greatly helps models to overcome the distraction issue.

As seen from lines 4 and 5, retrieval can still help our method to improve, though marginally from 48.1 to 48.4 on average, which is not as substantial as the improvement seen when applying retrieval to the vanilla-trained model (from 44.6 to 46.0 in lines 1-2).

\begin{figure}[!t]
\centering
\includegraphics[width=0.49\textwidth]{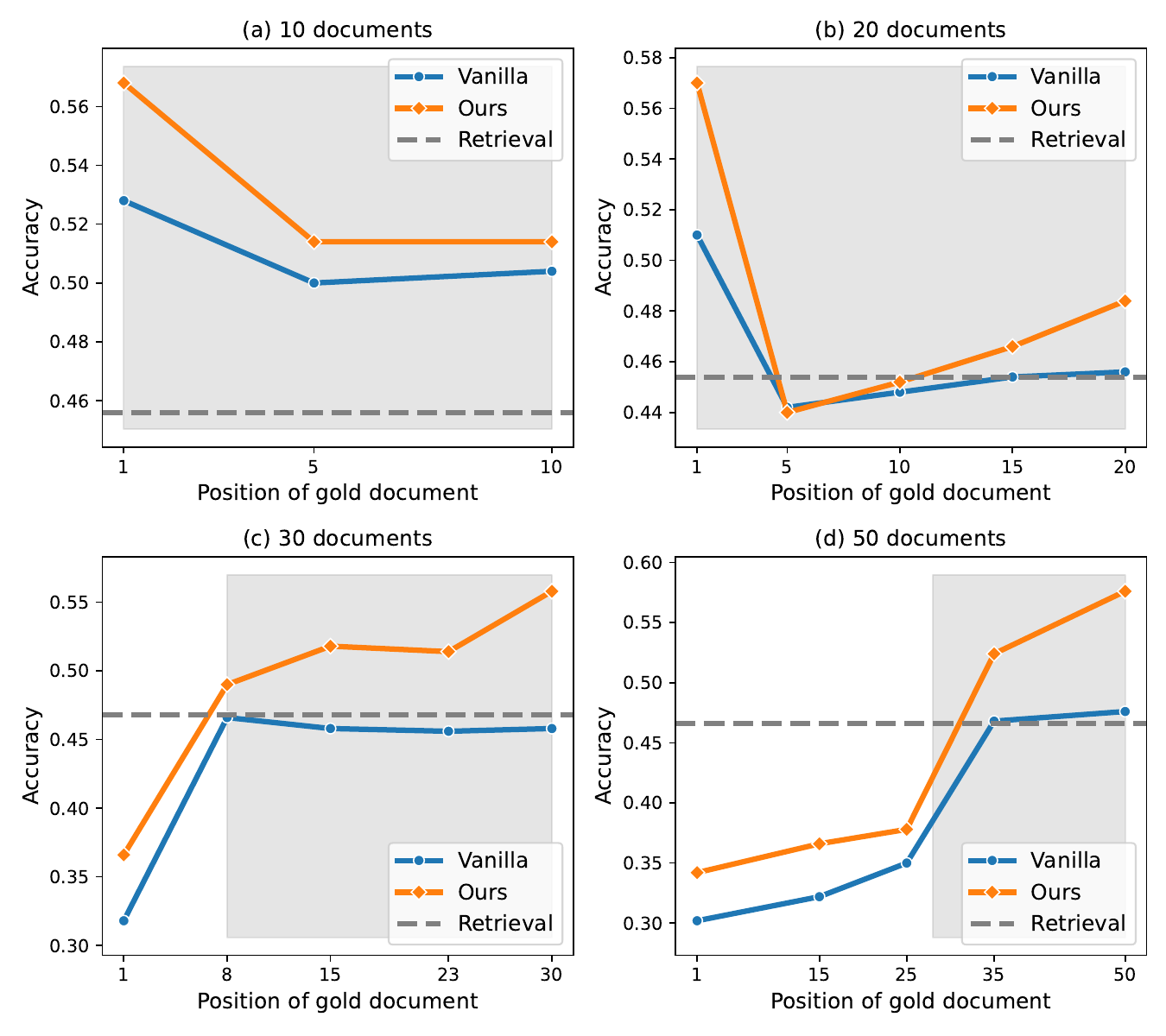}
\caption{Performance curves when placing the gold documents at different positions of the context at inference, when varying the total number of documents. The shaded area in each plot represents the last window context utilized by the sliding window attention mechanism in the \textit{Mistral} model.}
\label{fig: multidoc curve}
\end{figure}

\textbf{Utilizing the Inherent focus Bias.}
We vary the positions of the gold document in the input context using four different document lengths: 10, 20, 30, and 50, following \citet{liu2023lost}. The performance curves for different gold document positions are shown in Figure~\ref{fig: multidoc curve}. Our method consistently outperforms the vanilla-trained model across all positions. However, both models exhibit a focus bias, where their focus is not uniformly distributed within the context. This occurs despite a fine-tuning stage with randomly placed gold documents in a uniform distribution. This suggests that the focus bias is inherent and likely carried over from the model's pre-training stage and partially due the sliding window attention~\citep{child2019generating} being used. A deeper analysis is provided in Appendix~\ref{appendix: Positional Bias is Inherent}.

Figure~\ref{fig: multidoc curve} shows that, at certain positions, the performance curve surpasses that of using a retrieval model to filter relevant documents for input (indicated by gray dashed lines). At these positions, in-context retrieval outperforms the retrieval model. To leverage this positional bias, we explore an additional use of the retrieval model. Specifically, the retriever is employed to re-rank the documents based on their relevance scores with the question~\cite{liu-etal-2022-makes, xu2023retrieval}. 

More specifically, we arrange the documents from left to right, starting with the least relevant and progressing to the most relevant. This left-to-right arrangement is based on the observation that utilizing the external retrieval model is particularly beneficial in high-distraction scenarios, mitigating the ``lost in the beginning'' effect. We applied this re-ranking method to both models. As shown in lines 3 and 6 of Table~\ref{tab:main nqd results}, this re-ranking method generally improves both models (from 44.6 to 46.9 for the vanilla-trained model and from 48.1 to 52.0 for our method), providing a more substantial improvement than merely using retrieval to filter relevant documents.

\subsection{Analysis}

\begin{table}[!t]
\centering
\resizebox{0.42\textwidth}{!}{
\begin{tabular}{c l c c c c c} 
\toprule
\# & Methods & Qasper (F1) & NQd (EM) \\
\midrule
1 & Ours & \textbf{59.62} & \textbf{43.4}\\
2 & \, - Contra & 59.53 & 39.0 \\
3 & \, - Contra - DA & 48.65 & 40.0 \\
\midrule
4 & \, - Masking & 47.07 & 40.8 \\
\bottomrule
\end{tabular}}
\caption{Ablation on our proposed training data augmentation, masking and contrastive learning.}
\label{tab: ablation contra loss}
\end{table}

\textbf{Effectiveness of Our Proposed Components.}
We ablated the three components of our proposed method: data augmentation (DA), the contrastive learning objective (Contra), and the masking strategy (Masking) for irrelevant content in our data augmentation. Notably, removing both DA and Contra degrades our method to the vanilla training setting.
We considered the Qasper and NQd datasets, which provide gold evidence, ensuring high-quality augmented samples. The results of our ablation study are shown in Table~\ref{tab: ablation contra loss}.

We start with the performance of our full method (line 1), which is expected to have the highest performance among all ablated methods. 
When removing the contrastive learning objective (line 2), interestingly, the performance from Qasper does not decline much, but there is a huge decline in NQd that is even lower than the vanilla training model (line 3). We argue that, since the distractors in multi-document NQd are completely unrelated, contrastive learning plays a crucial role in distinguishing the augmented samples only with gold documents from the original ones where the gold documents are hidden.
Conversely, for the single-document Qasper, the model effectively learns on Data Augmentation (DA) because of the coherent original context.
Removing the masking strategy (line 4) also significantly impacts performance. Without masking, the model's performance on Qasper drops below the vanilla training method, while it marginally improves on NQd. This indicates that the masking token helps the model learn to ignore irrelevant information in Qasper, while in NQd, irrelevant information can be fully ignored without it.

\textbf{Case study.} We present a case study using the attention mechanism of our trained model to demonstrate that our method effectively teaches the model to focus on relevant content within a long input context. Detailed results and analysis are provided in Appendix~\ref{appendix: case study}.

\section{Conclusion}
In this work, we address the distraction issue in long-context LLMs by introducing a novel training method anchored on two key techniques: retrieval-based data augmentation and contrastive learning. Our method implicitly guides the LLM during fine-tuning to focus on the relevant information within lengthy contexts, thereby enhancing its ability to effectively utilize the long context. Extensive experiments on both single-document and multi-document benchmarks have demonstrated the effectiveness of the proposed method, outperforming baselines with just a few hundred fine-tuning steps. 

One possible future direction is to apply our method to a broader range of applications, particularly where identifying important chunks is challenging for a retriever model.

\section{Limitations}
While our method has demonstrated effectiveness in the QA task, which is a significant application area, there are several limitations to consider. 

The hypothesis that focusing only on the sub-context is sufficient to answer questions remains untested in other domains such as long-context summarization, which requires further investigation. Additionally, the effectiveness of our approach partially depends on the quality of the retriever; poor performance by the retriever could diminish the benefits. Lastly, the positional bias inherent in the model's architecture and pre-training stage can influence performance, suggesting the need for future work to mitigate this bias. 

\bibliography{custom}

\begin{thebibliography}{40}
\providecommand{\natexlab}[1]{#1}

\bibitem[{An et~al.(2022)An, Feng, Lv, Kong, Qiu, and Huang}]{an2022cont}
Chenxin An, Jiangtao Feng, Kai Lv, Lingpeng Kong, Xipeng Qiu, and Xuanjing Huang. 2022.
\newblock \href {https://openreview.net/forum?id=mjVZw5ADSbX} {Co{NT}: Contrastive neural text generation}.
\newblock In \emph{Advances in Neural Information Processing Systems}.

\bibitem[{Asai et~al.(2023)Asai, Wu, Wang, Sil, and Hajishirzi}]{asai2023self}
Akari Asai, Zeqiu Wu, Yizhong Wang, Avirup Sil, and Hannaneh Hajishirzi. 2023.
\newblock Self-rag: Learning to retrieve, generate, and critique through self-reflection.
\newblock \emph{arXiv preprint arXiv:2310.11511}.

\bibitem[{Brown et~al.(2020)Brown, Mann, Ryder, Subbiah, Kaplan, Dhariwal, Neelakantan, Shyam, Sastry, Askell, Agarwal, Herbert-Voss, Krueger, Henighan, Child, Ramesh, Ziegler, Wu, Winter, Hesse, Chen, Sigler, Litwin, Gray, Chess, Clark, Berner, McCandlish, Radford, Sutskever, and Amodei}]{NEURIPS2020_1457c0d6}
Tom Brown, Benjamin Mann, Nick Ryder, Melanie Subbiah, Jared~D Kaplan, Prafulla Dhariwal, Arvind Neelakantan, Pranav Shyam, Girish Sastry, Amanda Askell, Sandhini Agarwal, Ariel Herbert-Voss, Gretchen Krueger, Tom Henighan, Rewon Child, Aditya Ramesh, Daniel Ziegler, Jeffrey Wu, Clemens Winter, Chris Hesse, Mark Chen, Eric Sigler, Mateusz Litwin, Scott Gray, Benjamin Chess, Jack Clark, Christopher Berner, Sam McCandlish, Alec Radford, Ilya Sutskever, and Dario Amodei. 2020.
\newblock \href {https://proceedings.neurips.cc/paper_files/paper/2020/file/1457c0d6bfcb4967418bfb8ac142f64a-Paper.pdf} {Language models are few-shot learners}.
\newblock In \emph{Advances in Neural Information Processing Systems}, volume~33, pages 1877--1901.

\bibitem[{Caciularu et~al.(2022)Caciularu, Dagan, Goldberger, and Cohan}]{caciularu-etal-2022-long}
Avi Caciularu, Ido Dagan, Jacob Goldberger, and Arman Cohan. 2022.
\newblock \href {https://aclanthology.org/2022.naacl-main.207} {Long context question answering via supervised contrastive learning}.
\newblock In \emph{Proceedings of the 2022 Conference of the North American Chapter of the Association for Computational Linguistics: Human Language Technologies}, pages 2872--2879.

\bibitem[{Chen et~al.(2023{\natexlab{a}})Chen, Wong, Chen, and Tian}]{chen2023extending}
Shouyuan Chen, Sherman Wong, Liangjian Chen, and Yuandong Tian. 2023{\natexlab{a}}.
\newblock Extending context window of large language models via positional interpolation.
\newblock \emph{arXiv preprint arXiv:2306.15595}.

\bibitem[{Chen et~al.(2020)Chen, Kornblith, Norouzi, and Hinton}]{pmlr-v119-chen20j}
Ting Chen, Simon Kornblith, Mohammad Norouzi, and Geoffrey Hinton. 2020.
\newblock \href {https://proceedings.mlr.press/v119/chen20j.html} {A simple framework for contrastive learning of visual representations}.
\newblock In \emph{Proceedings of the 37th International Conference on Machine Learning}, volume 119, pages 1597--1607.

\bibitem[{Chen et~al.(2023{\natexlab{b}})Chen, Qian, Tang, Lai, Liu, Han, and Jia}]{chen2023longlora}
Yukang Chen, Shengju Qian, Haotian Tang, Xin Lai, Zhijian Liu, Song Han, and Jiaya Jia. 2023{\natexlab{b}}.
\newblock Longlora: Efficient fine-tuning of long-context large language models.
\newblock \emph{arXiv preprint arXiv:2309.12307}.

\bibitem[{Child et~al.(2019)Child, Gray, Radford, and Sutskever}]{child2019generating}
Rewon Child, Scott Gray, Alec Radford, and Ilya Sutskever. 2019.
\newblock Generating long sequences with sparse transformers.
\newblock \emph{arXiv preprint arXiv:1904.10509}.

\bibitem[{Dasigi et~al.(2021)Dasigi, Lo, Beltagy, Cohan, Smith, and Gardner}]{dasigi-etal-2021-dataset}
Pradeep Dasigi, Kyle Lo, Iz~Beltagy, Arman Cohan, Noah~A. Smith, and Matt Gardner. 2021.
\newblock \href {https://aclanthology.org/2021.naacl-main.365} {A dataset of information-seeking questions and answers anchored in research papers}.
\newblock In \emph{Proceedings of the 2021 Conference of the North American Chapter of the Association for Computational Linguistics: Human Language Technologies}, pages 4599--4610.

\bibitem[{Ding et~al.(2023)Ding, Ma, Dong, Zhang, Huang, Wang, and Wei}]{ding2023longnet}
Jiayu Ding, Shuming Ma, Li~Dong, Xingxing Zhang, Shaohan Huang, Wenhui Wang, and Furu Wei. 2023.
\newblock Longnet: Scaling transformers to 1,000,000,000 tokens.
\newblock In \emph{Proceedings of the 10th International Conference on Learning Representations}.

\bibitem[{Fei et~al.(2023)Fei, Niu, Zhou, Hou, Bai, Deng, and Han}]{fei2023extending}
Weizhi Fei, Xueyan Niu, Pingyi Zhou, Lu~Hou, Bo~Bai, Lei Deng, and Wei Han. 2023.
\newblock Extending context window of large language models via semantic compression.
\newblock \emph{arXiv preprint arXiv:2312.09571}.

\bibitem[{Gao et~al.(2024)Gao, Xiong, Gao, Jia, Pan, Bi, Dai, Sun, Guo, Wang, and Wang}]{gao2024retrievalaugmented}
Yunfan Gao, Yun Xiong, Xinyu Gao, Kangxiang Jia, Jinliu Pan, Yuxi Bi, Yi~Dai, Jiawei Sun, Qianyu Guo, Meng Wang, and Haofen Wang. 2024.
\newblock Retrieval-augmented generation for large language models: A survey.
\newblock \emph{arXiv preprint arXiv:2312.10997}.

\bibitem[{Guu et~al.(2020)Guu, Lee, Tung, Pasupat, and Chang}]{pmlr-v119-guu20a}
Kelvin Guu, Kenton Lee, Zora Tung, Panupong Pasupat, and Mingwei Chang. 2020.
\newblock \href {https://proceedings.mlr.press/v119/guu20a.html} {Retrieval augmented language model pre-training}.
\newblock In \emph{Proceedings of the 37th International Conference on Machine Learning}, volume 119, pages 3929--3938.

\bibitem[{He et~al.(2023)He, Pan, Dong, Song, Liu, Liang, Wang, Sun, Zhang, Xie et~al.}]{he2023never}
Junqing He, Kunhao Pan, Xiaoqun Dong, Zhuoyang Song, Yibo Liu, Yuxin Liang, Hao Wang, Qianguo Sun, Songxin Zhang, Zejian Xie, et~al. 2023.
\newblock Never lost in the middle: Improving large language models via attention strengthening question answering.
\newblock \emph{arXiv e-prints}, pages arXiv--2311.

\bibitem[{Hu et~al.(2022)Hu, yelong shen, Wallis, Allen-Zhu, Li, Wang, Wang, and Chen}]{hu2022lora}
Edward~J Hu, yelong shen, Phillip Wallis, Zeyuan Allen-Zhu, Yuanzhi Li, Shean Wang, Lu~Wang, and Weizhu Chen. 2022.
\newblock \href {https://openreview.net/forum?id=nZeVKeeFYf9} {Lo{RA}: Low-rank adaptation of large language models}.
\newblock In \emph{International Conference on Learning Representations}.

\bibitem[{Izacard et~al.(2022)Izacard, Caron, Hosseini, Riedel, Bojanowski, Joulin, and Grave}]{izacard2022unsupervised}
Gautier Izacard, Mathilde Caron, Lucas Hosseini, Sebastian Riedel, Piotr Bojanowski, Armand Joulin, and Edouard Grave. 2022.
\newblock \href {https://openreview.net/forum?id=jKN1pXi7b0} {Unsupervised dense information retrieval with contrastive learning}.
\newblock \emph{Transactions on Machine Learning Research}.

\bibitem[{Jain et~al.(2023)Jain, Zhang, Ahmad, Wang, Nan, Li, Tan, Nallapati, Ray, Bhatia, Ma, and Xiang}]{jain-etal-2023-contraclm}
Nihal Jain, Dejiao Zhang, Wasi~Uddin Ahmad, Zijian Wang, Feng Nan, Xiaopeng Li, Ming Tan, Ramesh Nallapati, Baishakhi Ray, Parminder Bhatia, Xiaofei Ma, and Bing Xiang. 2023.
\newblock \href {https://aclanthology.org/2023.acl-long.355} {{C}ontra{CLM}: Contrastive learning for causal language model}.
\newblock In \emph{Proceedings of the 61st Annual Meeting of the Association for Computational Linguistics}, pages 6436--6459.

\bibitem[{Jiang et~al.(2023)Jiang, Sablayrolles, Mensch, Bamford, Chaplot, Casas, Bressand, Lengyel, Lample, Saulnier et~al.}]{jiang2023mistral}
Albert~Q Jiang, Alexandre Sablayrolles, Arthur Mensch, Chris Bamford, Devendra~Singh Chaplot, Diego de~las Casas, Florian Bressand, Gianna Lengyel, Guillaume Lample, Lucile Saulnier, et~al. 2023.
\newblock \href {https://arxiv.org/abs/2310.06825} {Mistral 7b}.
\newblock \emph{arXiv preprint arXiv:2310.06825}.

\bibitem[{Karpukhin et~al.(2020)Karpukhin, Oguz, Min, Lewis, Wu, Edunov, Chen, and Yih}]{karpukhin-etal-2020-dense}
Vladimir Karpukhin, Barlas Oguz, Sewon Min, Patrick Lewis, Ledell Wu, Sergey Edunov, Danqi Chen, and Wen-tau Yih. 2020.
\newblock \href {https://aclanthology.org/2020.emnlp-main.550} {Dense passage retrieval for open-domain question answering}.
\newblock In \emph{Proceedings of the 2020 Conference on Empirical Methods in Natural Language Processing (EMNLP)}.

\bibitem[{Kwiatkowski et~al.(2019)Kwiatkowski, Palomaki, Redfield, Collins, Parikh, Alberti, Epstein, Polosukhin, Devlin, Lee, Toutanova, Jones, Kelcey, Chang, Dai, Uszkoreit, Le, and Petrov}]{kwiatkowski-etal-2019-natural}
Tom Kwiatkowski, Jennimaria Palomaki, Olivia Redfield, Michael Collins, Ankur Parikh, Chris Alberti, Danielle Epstein, Illia Polosukhin, Jacob Devlin, Kenton Lee, Kristina Toutanova, Llion Jones, Matthew Kelcey, Ming-Wei Chang, Andrew~M. Dai, Jakob Uszkoreit, Quoc Le, and Slav Petrov. 2019.
\newblock \href {https://aclanthology.org/Q19-1026} {Natural questions: A benchmark for question answering research}.
\newblock \emph{Transactions of the Association for Computational Linguistics}, pages 452--466.

\bibitem[{Lee et~al.(2021)Lee, Lee, and Hwang}]{lee2021contrastive}
Seanie Lee, Dong~Bok Lee, and Sung~Ju Hwang. 2021.
\newblock \href {https://openreview.net/forum?id=Wga_hrCa3P3} {Contrastive learning with adversarial perturbations for conditional text generation}.
\newblock In \emph{International Conference on Learning Representations}.

\bibitem[{Lewis et~al.(2020)Lewis, Perez, Piktus, Petroni, Karpukhin, Goyal, K{\"u}ttler, Lewis, Yih, Rockt{\"a}schel et~al.}]{lewis2020retrieval}
Patrick Lewis, Ethan Perez, Aleksandra Piktus, Fabio Petroni, Vladimir Karpukhin, Naman Goyal, Heinrich K{\"u}ttler, Mike Lewis, Wen-tau Yih, Tim Rockt{\"a}schel, et~al. 2020.
\newblock \href {https://proceedings.neurips.cc/paper_files/paper/2020/file/6b493230205f780e1bc26945df7481e5-Paper.pdf} {Retrieval-augmented generation for knowledge-intensive nlp tasks}.
\newblock \emph{Advances in Neural Information Processing Systems}, 33:9459--9474.

\bibitem[{Li et~al.(2023)Li, Dong, Guerin, and Lin}]{li-etal-2023-compressing}
Yucheng Li, Bo~Dong, Frank Guerin, and Chenghua Lin. 2023.
\newblock \href {https://aclanthology.org/2023.emnlp-main.391} {Compressing context to enhance inference efficiency of large language models}.
\newblock In \emph{Proceedings of the 2023 Conference on Empirical Methods in Natural Language Processing}, pages 6342--6353.

\bibitem[{Liu et~al.(2022)Liu, Shen, Zhang, Dolan, Carin, and Chen}]{liu-etal-2022-makes}
Jiachang Liu, Dinghan Shen, Yizhe Zhang, Bill Dolan, Lawrence Carin, and Weizhu Chen. 2022.
\newblock \href {https://aclanthology.org/2022.deelio-1.10} {What makes good in-context examples for {GPT}-3?}
\newblock In \emph{Proceedings of Deep Learning Inside Out (DeeLIO 2022): The 3rd Workshop on Knowledge Extraction and Integration for Deep Learning Architectures}, pages 100--114.

\bibitem[{Liu et~al.(2023)Liu, Lin, Hewitt, Paranjape, Bevilacqua, Petroni, and Liang}]{liu2023lost}
Nelson~F Liu, Kevin Lin, John Hewitt, Ashwin Paranjape, Michele Bevilacqua, Fabio Petroni, and Percy Liang. 2023.
\newblock \href {https://arxiv.org/abs/2307.03172} {Lost in the middle: How language models use long contexts}.
\newblock \emph{arXiv preprint arXiv:2307.03172}.

\bibitem[{Loshchilov and Hutter(2019)}]{loshchilov2018decoupled}
Ilya Loshchilov and Frank Hutter. 2019.
\newblock \href {https://openreview.net/forum?id=Bkg6RiCqY7} {Decoupled weight decay regularization}.
\newblock In \emph{International Conference on Learning Representations}.

\bibitem[{Pang et~al.(2022)Pang, Parrish, Joshi, Nangia, Phang, Chen, Padmakumar, Ma, Thompson, He, and Bowman}]{pang-etal-2022-quality}
Richard~Yuanzhe Pang, Alicia Parrish, Nitish Joshi, Nikita Nangia, Jason Phang, Angelica Chen, Vishakh Padmakumar, Johnny Ma, Jana Thompson, He~He, and Samuel Bowman. 2022.
\newblock \href {https://aclanthology.org/2022.naacl-main.391} {{Q}u{ALITY}: Question answering with long input texts, yes!}
\newblock In \emph{Proceedings of the 2022 Conference of the North American Chapter of the Association for Computational Linguistics: Human Language Technologies}, pages 5336--5358.

\bibitem[{Press et~al.(2021)Press, Smith, and Lewis}]{press2021train}
Ofir Press, Noah~A Smith, and Mike Lewis. 2021.
\newblock Train short, test long: Attention with linear biases enables input length extrapolation.
\newblock \emph{arXiv preprint arXiv:2108.12409}.

\bibitem[{Radford et~al.(2021{\natexlab{a}})Radford, Kim, Hallacy, Ramesh, Goh, Agarwal, Sastry, Askell, Mishkin, Clark, Krueger, and Sutskever}]{pmlr-v139-radford21a}
Alec Radford, Jong~Wook Kim, Chris Hallacy, Aditya Ramesh, Gabriel Goh, Sandhini Agarwal, Girish Sastry, Amanda Askell, Pamela Mishkin, Jack Clark, Gretchen Krueger, and Ilya Sutskever. 2021{\natexlab{a}}.
\newblock \href {https://proceedings.mlr.press/v139/radford21a.html} {Learning transferable visual models from natural language supervision}.
\newblock In \emph{Proceedings of the 38th International Conference on Machine Learning}, pages 8748--8763.

\bibitem[{Radford et~al.(2021{\natexlab{b}})Radford, Kim, Hallacy, Ramesh, Goh, Agarwal, Sastry, Askell, Mishkin, Clark et~al.}]{radford2021learning}
Alec Radford, Jong~Wook Kim, Chris Hallacy, Aditya Ramesh, Gabriel Goh, Sandhini Agarwal, Girish Sastry, Amanda Askell, Pamela Mishkin, Jack Clark, et~al. 2021{\natexlab{b}}.
\newblock \href {https://proceedings.mlr.press/v139/radford21a/radford21a.pdf} {Learning transferable visual models from natural language supervision}.
\newblock In \emph{International conference on machine learning}, pages 8748--8763. PMLR.

\bibitem[{Radford et~al.(2018)Radford, Narasimhan, Salimans, Sutskever et~al.}]{radford2018improving}
Alec Radford, Karthik Narasimhan, Tim Salimans, Ilya Sutskever, et~al. 2018.
\newblock \href {https://cdn.openai.com/research-covers/language-unsupervised/language_understanding_paper.pdf} {Improving language understanding by generative pre-training}.

\bibitem[{Shi et~al.(2023{\natexlab{a}})Shi, Chen, Misra, Scales, Dohan, Chi, Sch\"{a}rli, and Zhou}]{pmlr-v202-shi23a}
Freda Shi, Xinyun Chen, Kanishka Misra, Nathan Scales, David Dohan, Ed~H. Chi, Nathanael Sch\"{a}rli, and Denny Zhou. 2023{\natexlab{a}}.
\newblock \href {https://proceedings.mlr.press/v202/shi23a.html} {Large language models can be easily distracted by irrelevant context}.
\newblock In \emph{Proceedings of the 40th International Conference on Machine Learning}, pages 31210--31227.

\bibitem[{Shi et~al.(2023{\natexlab{b}})Shi, Min, Yasunaga, Seo, James, Lewis, Zettlemoyer, and Yih}]{shi2023replug}
Weijia Shi, Sewon Min, Michihiro Yasunaga, Minjoon Seo, Rich James, Mike Lewis, Luke Zettlemoyer, and Wen-tau Yih. 2023{\natexlab{b}}.
\newblock Replug: Retrieval-augmented black-box language models.
\newblock \emph{arXiv preprint arXiv:2301.12652}.

\bibitem[{Su et~al.(2024)Su, Ahmed, Lu, Pan, Bo, and Liu}]{su2024roformer}
Jianlin Su, Murtadha Ahmed, Yu~Lu, Shengfeng Pan, Wen Bo, and Yunfeng Liu. 2024.
\newblock \href {https://www.sciencedirect.com/science/article/pii/S0925231223011864?casa_token=J6LvMvJFf-YAAAAA:LgIVKfPq9m-iYrbW6paIBWU_lsM3i8LC9uoUBK8i8lmowhdzxExrsbfm1vy3U8pUVXMvKkMWkyU} {Roformer: Enhanced transformer with rotary position embedding}.
\newblock \emph{Neurocomputing}, 568:127063.

\bibitem[{Su et~al.(2022)Su, Lan, Wang, Yogatama, Kong, and Collier}]{su2022a}
Yixuan Su, Tian Lan, Yan Wang, Dani Yogatama, Lingpeng Kong, and Nigel Collier. 2022.
\newblock \href {https://openreview.net/forum?id=V88BafmH9Pj} {A contrastive framework for neural text generation}.
\newblock In \emph{Advances in Neural Information Processing Systems}.

\bibitem[{Vaswani et~al.(2017)Vaswani, Shazeer, Parmar, Uszkoreit, Jones, Gomez, Kaiser, and Polosukhin}]{NIPS2017_3f5ee243}
Ashish Vaswani, Noam Shazeer, Niki Parmar, Jakob Uszkoreit, Llion Jones, Aidan~N Gomez, \L~ukasz Kaiser, and Illia Polosukhin. 2017.
\newblock \href {https://proceedings.neurips.cc/paper_files/paper/2017/file/3f5ee243547dee91fbd053c1c4a845aa-Paper.pdf} {Attention is all you need}.
\newblock In \emph{Advances in Neural Information Processing Systems}. Curran Associates, Inc.

\bibitem[{Wei et~al.(2022)Wei, Wang, Schuurmans, Bosma, ichter, Xia, Chi, Le, and Zhou}]{NEURIPS2022_9d560961}
Jason Wei, Xuezhi Wang, Dale Schuurmans, Maarten Bosma, brian ichter, Fei Xia, Ed~Chi, Quoc~V Le, and Denny Zhou. 2022.
\newblock \href {https://proceedings.neurips.cc/paper_files/paper/2022/file/9d5609613524ecf4f15af0f7b31abca4-Paper-Conference.pdf} {Chain-of-thought prompting elicits reasoning in large language models}.
\newblock In \emph{Advances in Neural Information Processing Systems}, volume~35, pages 24824--24837.

\bibitem[{Xiong et~al.(2023)Xiong, Liu, Molybog, Zhang, Bhargava, Hou, Martin, Rungta, Sankararaman, Oguz et~al.}]{xiong2023effective}
Wenhan Xiong, Jingyu Liu, Igor Molybog, Hejia Zhang, Prajjwal Bhargava, Rui Hou, Louis Martin, Rashi Rungta, Karthik~Abinav Sankararaman, Barlas Oguz, et~al. 2023.
\newblock Effective long-context scaling of foundation models.
\newblock \emph{arXiv preprint arXiv:2309.16039}.

\bibitem[{Xu et~al.(2023)Xu, Ping, Wu, McAfee, Zhu, Liu, Subramanian, Bakhturina, Shoeybi, and Catanzaro}]{xu2023retrieval}
Peng Xu, Wei Ping, Xianchao Wu, Lawrence McAfee, Chen Zhu, Zihan Liu, Sandeep Subramanian, Evelina Bakhturina, Mohammad Shoeybi, and Bryan Catanzaro. 2023.
\newblock \href {https://arxiv.org/abs/2310.03025} {Retrieval meets long context large language models}.
\newblock \emph{arXiv preprint arXiv:2310.03025}.

\bibitem[{Zhang et~al.(2020)Zhang, Zhao, Saleh, and Liu}]{zhang2020pegasus}
Jingqing Zhang, Yao Zhao, Mohammad Saleh, and Peter Liu. 2020.
\newblock \href {https://proceedings.mlr.press/v119/zhang20ae/zhang20ae.pdf} {Pegasus: Pre-training with extracted gap-sentences for abstractive summarization}.
\newblock In \emph{International Conference on Machine Learning}, pages 11328--11339. PMLR.

\end{thebibliography}

\appendix

\section{Implementation Details}
\begin{table}[!t]
\centering
\caption{Details of the datasets.}
\label{tab:dataset stats}
\resizebox{0.48\textwidth}{!}{
\begin{tabular}{l  c c  c } 
\toprule
\textbf{Datasets} & \textbf{\#Training} & \textbf{\#Test} & \textbf{Avg. context length} \\
\midrule
Qasper & 2567 & 1005 & 5913 \\
QuALITY & 2523 & 2086 & 7190\\
NQd & 1500 & 500 & 8163 \\
\bottomrule
\end{tabular}}
\end{table}

\subsection{Datasets}\label{appendix: datasets}
The details of those datasets are shown in Table~\ref{tab:dataset stats}, where we report the size of the training and test set, as well as the average token length of the context of the training samples.
For each sample, we used the template ``Article:\{Context\}\textbackslash n Question: \{Question\}\textbackslash n Answer:'' to format the prompt. At training, the gold answer is appended to the prompt, where the model learns the whole sequences for both CLM and contrastive learning objectives. Whereas at inference, only the prompt is given to the model to generate an answer for evaluation.

\subsection{Training Procedure}\label{appendix: training procedure}
We fine-tuned the \textit{Mistral}-7B model with LoRA on the synthetic natural questions training set with 100 steps, and 500 steps on both Qasper and QuALITY datasets. Due to the long context, each GPU can only fit one sample, and therefore the batch size is 8 on our 8-GPU setup. We use AdamW~\citep{loshchilov2018decoupled} as the optimizer, with with $\beta1 = 0.9$ and $\beta2 = 0.999$. The learning rate was set at 2e-5. Additionally, we implemented a linear learning rate warmup strategy for the initial 5\% of the training steps.

\section{Analysis of Focus Bias} \label{appendix: Positional Bias is Inherent}
\begin{table}[!t]
\centering
\resizebox{0.49\textwidth}{!}{
\begin{tabular}{l c c c c c} 
\toprule
\multirow{2}{*}{\#Documents}& \multicolumn{2}{c}{Group 1} & \multicolumn{3}{c}{Group 2} \\
\cmidrule(lr){2-3} \cmidrule(lr){4-6}
 & 10 & 20 & 30 & 40 & 50 \\
\midrule
Avg. Context Length & 1.6k & 3.3k & 4.9k & 6.5k & 8.2k \\
\bottomrule
\end{tabular}}
\caption{Average context length (in tokens) of the NQd test sample. Groups are decided by whether the input context can fit into a single attention window with 4k tokens.}
\label{tab: NQd context length}
\end{table}

As shown in Figure~\ref{fig: multidoc curve}, when the document lengths are limited to 10 and 20, there is a ``lost in the middle'' phenomenon, which is consistent with the findings from \citet{liu2023lost}. Conversely, when the document lengths exceed 30, there is a ``lost in the beginning'' trend. The difference may stem from the use of sliding window attention~\citep{child2019generating} in the \textit{Mistral} model, a technique to save memory for the quadratic nature of self-attention.

The context lengths from different document lengths are shown in Table~\ref{tab: NQd context length}, which can be divided into two groups based on whether an attention window can fit the whole input context (the window size of \textit{Mistral} is 4k). In the case of inference with 10 and 20 documents, the tokens of the input context are always within the same attention window. Therefore, the question always has direct attention to any position of the input context, matching the settings of \citet{liu2023lost}.

On the other hand, the beginning of the context from 30 and 50 documents is always outside the last attention window where the questions reside. Therefore, the attention between the gold document at the beginning and the question at the end is achieved by connecting two windows at different layers of transformers~\cite{child2019generating}. This results in the observed ``lost in the beginning'' pattern.

\section{Case Study} \label{appendix: case study}
\begin{figure*}[!t]
\centering
\includegraphics[width=0.95\textwidth]{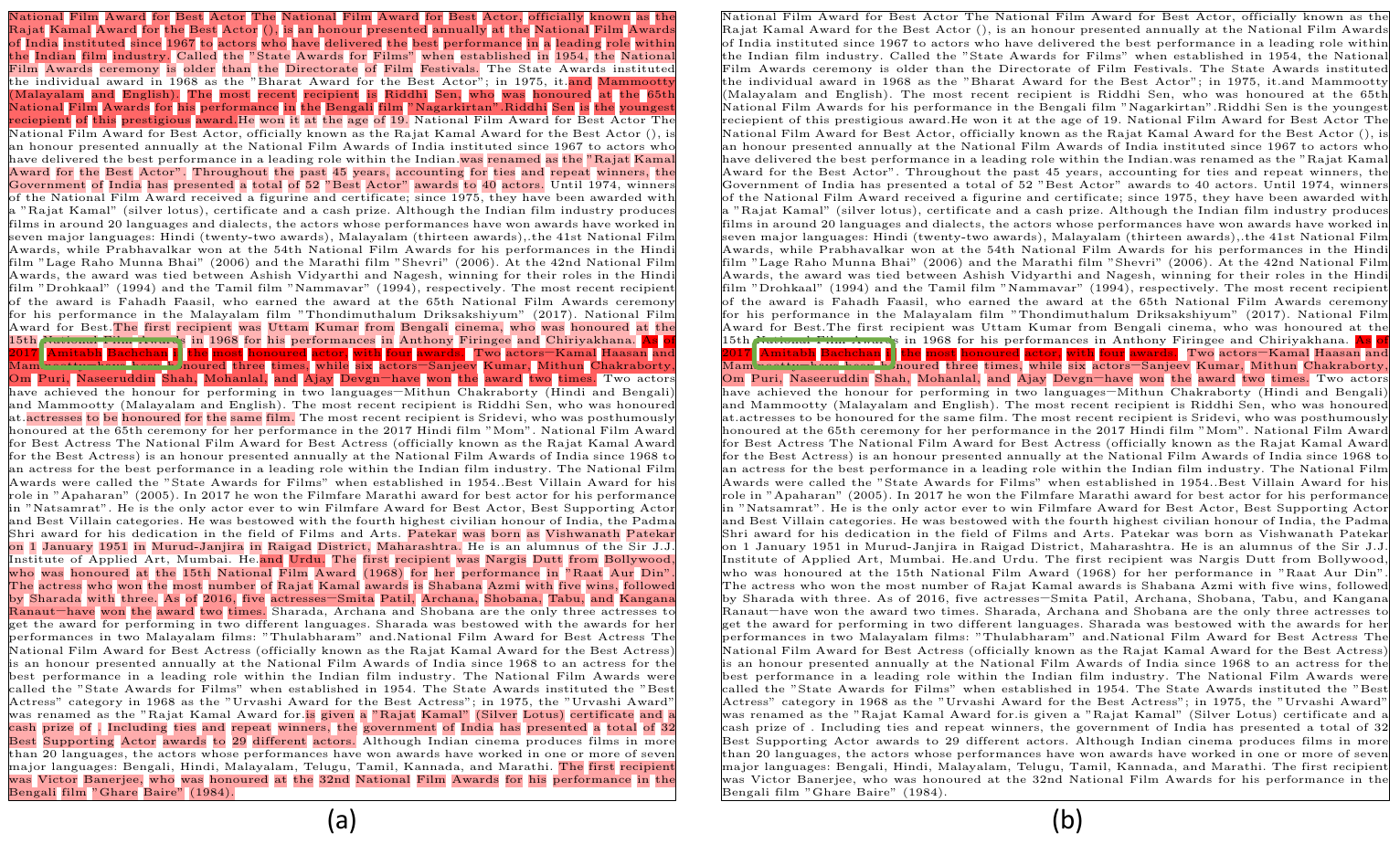}
\caption{The sentence-level attention maps between the question ``Which Indian actor has won most national awards?'' and a concatenation of 10 documents. (a) is the vanilla method attention maps, and (b) is the re-focused attention maps after training with our method. The green rectangle is the location of the answer.}
\label{fig: attention case study}
\end{figure*}

We randomly select a test sample from our NQd dataset. To better visualize the attention heatmap, our sample only contains 10 documents where 9 are distractors and the gold one is in the middle. Specifically, we chose the attention scores of the last token before generation (``:'' from our template).
We averaged the attention score at the sentence level and compared the attention heatmap with or without training from our method.

As shown in Figure~\ref{fig: attention case study}, our method helps the model better focus on the true relevant content, or gold document. For the vanilla-trained model, its attention is widely spread. It attends to the distractors, which is unexpected. However, our method only attends to the gold document, showing the efficacy of the focus.

\end{document}